\documentclass[conference]{IEEEtran}

\IEEEoverridecommandlockouts
\usepackage{caption}
\usepackage{threeparttable}
\usepackage{tablefootnote}
\usepackage{varwidth}
\usepackage{amsmath}
\usepackage{xcolor} 
\usepackage{hyperref}
\usepackage{graphicx,epstopdf,algpseudocode,caption,url}   
\usepackage{multirow} 
\graphicspath{{./images}}
\usepackage{arydshln} 
\begin{document}

\title{Model-as-a-Service (MaaS): A Survey}  

\author{Wensheng Gan$ ^{1,2}$, Shicheng Wan$ ^{3*}$\thanks{\IEEEauthorrefmark{1}Corresponding author.}, Philip S. Yu$ ^{4}$  \\ \\
	$ ^{1} $Jinan University, Guangzhou 510632, China \\
	$ ^{2} $Pazhou Lab, Guangzhou 510330, China\\
        $ ^{3} $South China University of Technology, Guangzhou 510641, China.\\
        $ ^{4} $University of Illinois Chicago, Chicago, IL 60637, USA\\
        Email: \{wsgan001, scwan1998\}@gmail.com, psyu@uic.edu
}

\maketitle

\begin{abstract}
    Due to the increased number of parameters and data in the pre-trained model exceeding a certain level, a foundation model (e.g., a large language model) can significantly improve downstream task performance and emerge with some novel special abilities (e.g., deep learning, complex reasoning, and human alignment) that were not present before. Foundation models are a form of generative artificial intelligence (GenAI), and Model-as-a-Service (MaaS) has emerged as a groundbreaking paradigm that revolutionizes the deployment and utilization of GenAI models. MaaS represents a paradigm shift in how we use AI technologies and provides a scalable and accessible solution for developers and users to leverage pre-trained AI models without the need for extensive infrastructure or expertise in model training. In this paper, the introduction aims to provide a comprehensive overview of MaaS, its significance, and its implications for various industries. We provide a brief review of the development history of ``X-as-a-Service'' based on cloud computing and present the key technologies involved in MaaS. The development of GenAI models will become more democratized and flourish. We also review recent application studies of MaaS. Finally, we highlight several challenges and future issues in this promising area. MaaS is a new deployment and service paradigm for different AI-based models. We hope this review will inspire future research in the field of MaaS.
\end{abstract}


\IEEEpeerreviewmaketitle

\section{Introduction}  \label{sec:Introduction}

We currently live in an era of fast-developing within big data \cite{gan2017data,sun2022big}, artificial intelligence (AI) \cite{minsky1961steps}, and Web 3.0 \cite{gan2023web,wan2023web3}, organizations within various industries are increasingly leveraging the power of machine learning (ML) models \cite{jordan2015machine} to gain insights, automate processes, and make data-driven decisions. However, developing, training, and deploying ML models are challenging and resource-intensive. Especially in the digital age, generative artificial intelligence, e.g., large language model (LLM) \cite{hu2023survey}, has recently not only become a public-familiar word on social media but also a hot spot in the academic field. Generative artificial intelligence (GenAI) \cite{gozalo2023survey} refers to the ability of a machine learning model to generate new data that is similar to the training data. This is in contrast to the traditional AI approach, which is designed to recognize differences between distinct classes or categories of data. GenAI models are often adopted in tasks such as image generation, text generation \cite{raffel2020exploring}, and music composition \cite{epstein2023art}. LLM is an advanced model within generative AI that aims to simulate human language ability and intelligence (i.e., generation capability). Developers and engineers train LLM using massive amounts of textual data and then utilize deep learning algorithms \cite{vaswani2017attention,devlin2018bert,radford2019better} to understand as many languages as possible. Indeed, the training process is the core of LLM \cite{carlini2021extracting,zeng2023distributed}. It involves training the model on huge amounts of data to learn the statistical patterns and semantic relationships of language. During the training process \cite{zeng2023distributed}, LLM predicts the next word or sentence to optimize its parameters and improve its predictive abilities. Through iterative training, LLM gradually enhances its language generation and understanding quality. Benefiting from generating human language text capability, LLM has become an important service in various domains. It provides people with a lot of useful unexpected functions and outstanding applications, such as data analytic \cite{gan2019survey,sun2023internet}, natural language processing (NLP) \cite{thirunavukarasu2023large}, intelligent conversation system \cite{yu2022xdai}, machine translation \cite{garcia2023unreasonable}, information retrieval \cite{bonifacio2022inpars}, and text-to-multimodal \cite{yu2022commercemm}.

Since the foundation model is still in its exploration stage, there is no clear definition and consensus in academia and industry \cite{huang2023chatgpt,shen2023chatgpt}. This is where Model-as-a-Service (MaaS) comes into play. MaaS is a cloud computing-based service framework that offers AI and ML models and related infrastructure as a service to developers and businesses. It provides a convenient and cost-effective way to access and utilize large models without the need for extensive knowledge or infrastructure. MaaS allows users to leverage pre-trained ML models and algorithms through simple interfaces, application programming interfaces (APIs), or software development kits (SDKs). MaaS allows users to access functions of the large model through calling API without the need to train and maintain complex models themselves.

In simple terms, MaaS is a new business model. As the foundation infrastructure of the AI era, MaaS provides secure, efficient, and cost-effective model usage and development support for downstream applications. Looking at the entire industry structure of MaaS, the core idea follows the path of ``Model–Single Point Tool-Application Scenarios''. Users can directly call, develop, and deploy models in the cloud without the need to invest in building and maintaining the infrastructure, hardware, and specialized knowledge required for their own models. Large models, as a critical component of MaaS, are a product of the combination of ``high computing power and strong algorithms'' and will be a development trend for the future of artificial intelligence. Large models are rapidly evolving, transitioning AI from a ``craft workshop'' to a ``factory model'', achieving greater versatility and intelligence, and enabling AI to empower applications across various industries more widely. MaaS finds wide applications across multiple domains. In intelligent conversational systems, MaaS can be used to build chatbots, voice assistants, and other interactive systems that engage in smooth and natural conversations with users. In information retrieval, MaaS can provide intelligent search and recommendation functionalities, helping users quickly find the desired information. However, the development of MaaS also faces some challenges. Firstly, there are performance and stability considerations. Large models require significant training data and computational resources to achieve optimal performance, and they need to address limitations in tasks such as long-text generation and logical reasoning. Secondly, there are privacy and security concerns as MaaS may involve users' personal information and sensitive data, requiring appropriate security measures in data transmission and privacy protection.

In the growing landscape of AI and ML, MaaS is emerging as a groundbreaking paradigm that revolutionizes the deployment and utilization of AI models. MaaS provides a scalable and accessible solution for organizations and developers to leverage pre-trained AI models without the need for extensive infrastructure or expertise in model training. However, there has not yet been a comprehensive overview of MaaS proposed in the academic. This paper aims to introduce MaaS by highlighting key components of the MaaS model, their underlying technologies, the advantages of MaaS, and the differences from previous cloud computing-based service models.

\textbf{Contributions}: To fill this research gap, this paper tries to provide a comprehensive literature survey of MaaS, and the contributions are as follows:

\begin{itemize}    
    \item This is the first review to introduce the characteristics of the interaction and related technologies of MaaS, and we also review the history of cloud computing services.

    \item We analyze differences between traditional and GenAI model-based technology stacks and conclude the latter stack presents many new functions and features.

    \item We also highlight some promising application fields within MaaS that are currently developing or will be popular in the near future.
    
    \item We analyze the challenges and issues encountered by MaaS in more detail, including utilization constraints, model interpretability, technological ethics, and morality.
\end{itemize}

\textbf{Roadmap}: We first introduce the history of previous cloud-based service platforms in Section \ref{sec:History}. We analyzed the relationship between MaaS and previous XaaS in Section \ref{sec:Relationship}, and listed the benefits of MaaS in Section \ref{sec:Advantages}. We then introduce some key technologies related to MaaS in Section \ref{sec:Technologies} and present several case applications in Section \ref{sec:Applications}. We also discuss several challenges and issues faced by MaaS in Section \ref{sec:Challenges}. Finally, Section \ref{sec:Conclusion} concludes this survey with a discussion of potential future research. Note that Table \ref{Symbols} presents some basic symbols in this survey.

\begin{table}[h]
    \centering
    \caption{Summary of symbols and their explanations}
    \label{Symbols}
    \begin{tabular}{|c|c|}
        \hline
        \textbf{Symbol} & \textbf{Definition}                \\ \hline
        AI              & Artificial intelligence            \\ \hline
        GenAI           & Generative artificial intelligence \\ \hline
        ML &  Machine learning \\ \hline
        LLM             & Large language model              \\ \hline
        NLP             & Natural language processing       \\ \hline
        API             & Application programming interface \\ \hline
        	
        SaaS            & Software-as-a-Service             \\ \hline
        PaaS            & Platform-as-a-Service             \\ \hline
        IaaS            & Infrastructure-as-a-Service       \\ \hline
        MaaS            & Model-as-a-Service                \\ \hline
    \end{tabular}
\end{table}

\section{History of XaaS} \label{sec:History}

When reviewing the development history of Web 2.0, the ``X-as-a-Service'' model paradigm (XaaS) is typical of the second wave \cite{mohammed2021sufficient}. Indeed, since the complex and resource-intensive disadvantages of developing environment construction, the core target of XaaS is to provide users with infrastructure and then let them focus on solution design.

Software-as-a-Service (SaaS) \cite{cusumano2010cloud} is one of the earliest cloud computing service models, which originally emerged between the late 1990s and early 2000s. SaaS delivers internet-based applications and then allows users to use these applications according to the online subscription method. This new kind of Internet service lets users no longer need to install local software, such as BaiduNetdisk, Dropbox, and Gmail. In general, SaaS applications are capable of holding abundant users because of their good scalability. The SaaS provider handles almost all maintenance tasks, including updates, bug fixes, security, and data backups, while users are only permitted to do human-software interaction. However, since the number of users involved in the Internet world is continuously rising, the rise of personalized customization for different users (i.e., more software products in various fields) has led to an explosion of requirements and then formed an urgent issue.

Platform-as-a-Service (PaaS) \cite{van2018serverless} emerged after SaaS (around the mid-2000s). It provides program developers with a stable cloud computing platform for building, testing, and deploying applications. In PaaS solutions, a provider hosts both the hardware and software on their dedicated infrastructure and delivers this platform to users as an all-in-one solution, software stack, or service accessible through the Internet. Developers can primarily concentrate on coding, relieving them of the concerns related to maintaining and managing the foundational infrastructure or platform traditionally linked to the development process.  This paves the path for continued development and innovation with minimal interruptions, simultaneously diminishing the necessity for extensive infrastructure setup and coding efforts. Google's App Engine application platform \cite{chohan2010appscale,laxmaiah2019comparative} and Microsoft's Azure app service \cite{chappell2008introducing} are two classic examples of PaaS. PaaS providers are responsible for maintaining the underlying infrastructure, while users only focus on application development and management. Nevertheless, because of operational costs, PaaS may impose limitations on the different resources required for applications and workloads (e.g., computational resources, storage resources, and network resources). In some cases, specific application requirements might not be compatible with the PaaS environment, which could result in issues when deploying and running applications.

\begin{figure}[!ht]
    \centering
    \includegraphics[scale=0.28]{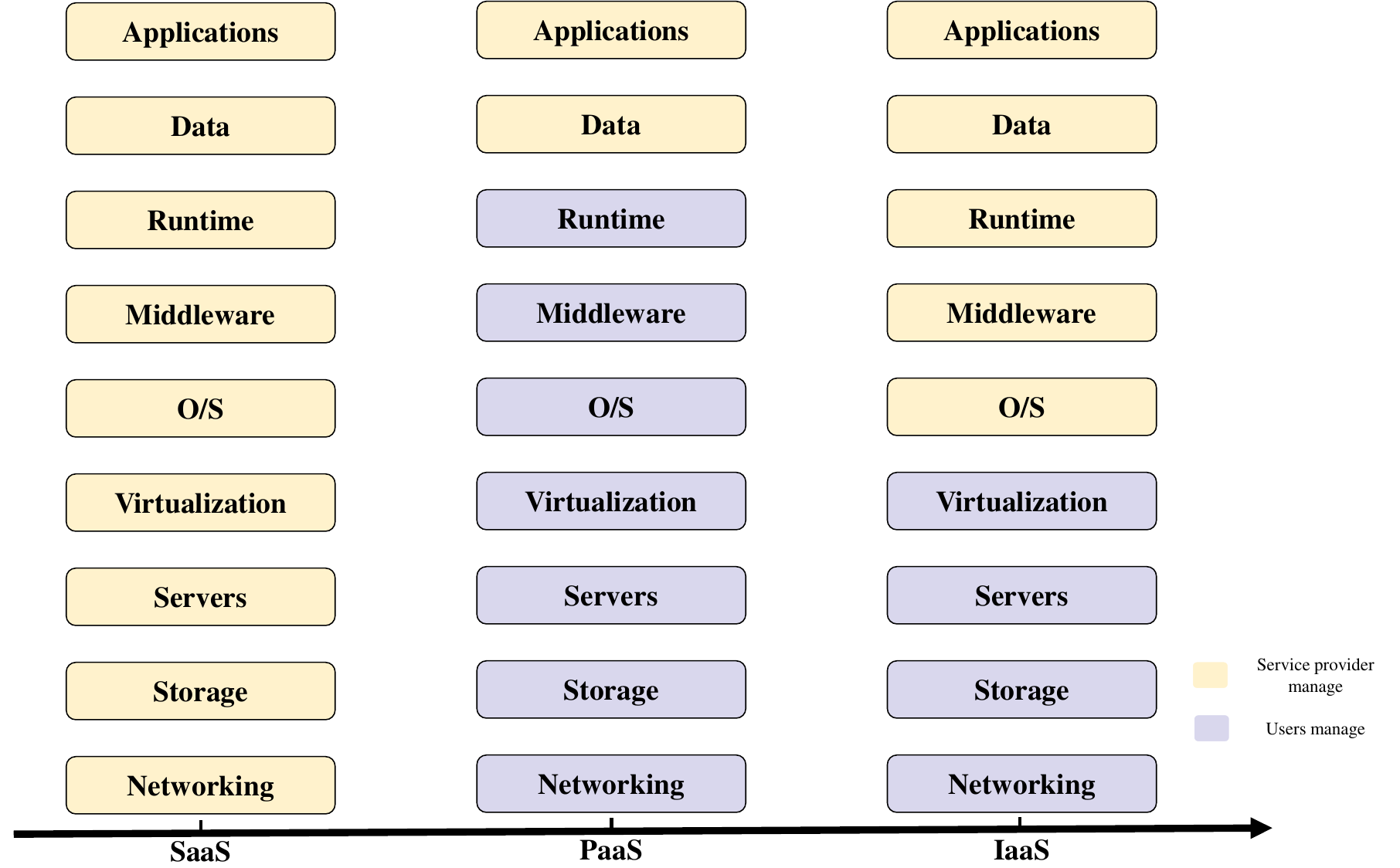}
    \caption{A comparison between the infrastructures involved by the user and service provider. (Source: \url{https://www.redhat.com/en/topics/cloud-computing/what-is-paas})}
    \label{fig:history}
\end{figure}

To provide users with greater flexibility, control, and adaptability to meet various application requirements and business scenarios, the latest model in cloud computing services, i.e., Infrastructure-as-a-Service (IaaS) \cite{manvi2014resource}, emerged around 2010. IaaS provides virtualized computing resources through web services, which allow users to rent infrastructure services such as servers, networking, and storage on-demand. Generally, users can utilize the infrastructure through application programming interfaces (APIs). Users obtain greater control over the configuration of fundamental infrastructure (e.g., virtual servers, storage, and networking) and take over responsibilities previously managed and maintained by PaaS providers within the IaaS model. The IaaS providers only focus on managing hardware modules and dealing with hardware issues. Engineers have developed many mature IaaS platforms (e.g., Google Compute Engine \cite{krishnan2015google}, DigitalOcean \cite{pikkuhookana2023software}, and Amazon Elastic Compute Cloud \cite{cloud2009amazon}). These platforms have been deeply integrated into our daily lives. Yet, SaaS offers broader applications to users and various non-technical departments, whereas PaaS and IaaS predominantly cater to development teams. Fig. \ref{fig:history} depicts the change path of XaaS.

\section{Relationship Between other XaaS} \label{sec:Relationship}

Similar to the previous XaaS models, in the forthcoming Web 3.0 era, MaaS is a new service model in the field of AI that involves providing access to pre-trained machine learning models as a service on the Internet. It involves hosting, managing, and giving developers access to models that have already been trained through APIs. This lets developers add AI functions to their own systems and apps. The target behind MaaS is to provide a platform where developers can access pre-trained machine learning models that have been trained on large datasets and optimized for specific tasks. Thus, MaaS abstracts away the complexities of model training and deployment and makes it easier for developers to leverage AI technologies. Besides, MaaS enables developers to focus on their specific use cases and application logic, without the need to delve into the intricacies of model training, hyperparameter tuning, and infrastructure management. By using MaaS, developers can save time and resources by utilizing existing models rather than building and training models from scratch. Additionally, in order to create sophisticated models and AI capabilities, it democratizes AI for a wider user base, including developers with no prior AI knowledge.

During the Web 2.0 age, SaaS, PaaS, and IaaS are all cloud computing-based solutions \cite{peng2009comparison}. They offer services over the Internet to reduce the need for on-premises deployment. In terms of cost consideration, this kind of subscription-based flexible payment model benefits both customers and providers significantly \cite{wang2005subscription}: 1) Customers do not consider how to invest in and maintain fundamental infrastructure, such as software and hardware, anymore. 2) Pieces of equipment procured by providers can be efficiently reused across customers, which results in substantial savings in acquisition expenses. 3) The high scalability of these models empowers customers to adapt resource usage and payment obligations according to their specific requirements. However, compared to traditional cloud service models, MaaS has some characteristics and advantages.

\begin{itemize}
    \item MaaS takes further consideration in the service model by focusing on the training, deployment, and invocation of AI and ML models. In contrast, IaaS provides basic computing resources; PaaS offers application development and deployment platforms; and SaaS provides complete application software. In particular, MaaS abstracts a higher level of service that enables users to directly utilize the capabilities of AI and ML models.

    \item MaaS aims to offer specialized services and customized functions for users to complete different tasks (such as conversation, video production, and artistic image creation). It is supposed to be a more professional tool for users in the field of development or application. Hence, the core components of MaaS are the training, deployment, and invocation of various AI models. Traditional cloud service models, however, focus more on providing infrastructure, platforms, or application software and offer lower levels of model selection and customization.

    \item MaaS provides a lot of simplified tools that make it easier for users to utilize and deploy AI models. In contrast, traditional cloud service models often require users to have more technical knowledge and experience to fully utilize the related services, and they also limit the widespread adoption of AI technologies. These can be a barrier for non-experts or most developers. MaaS lets AI usage be more democratic by providing user-friendly operation interfaces and low entry barriers.
\end{itemize}

In conclusion, MaaS refers to the delivery of machine learning models as a cloud-based service. As a service model specifically targeting machine learning models, MaaS differs from traditional cloud service models (i.e., IaaS, PaaS, and SaaS) in terms of abstraction level, service objects, model selection and customization, and usability. The advantages of MaaS are that it offers more advanced services and abstractions tailored to machine learning models. It is convenient and efficient for users to use and deploy machine learning models.

\section{Related Technologies of MaaS} \label{sec:Technologies}

\begin{figure}[!ht]
    \centering
    \includegraphics[scale=0.26]{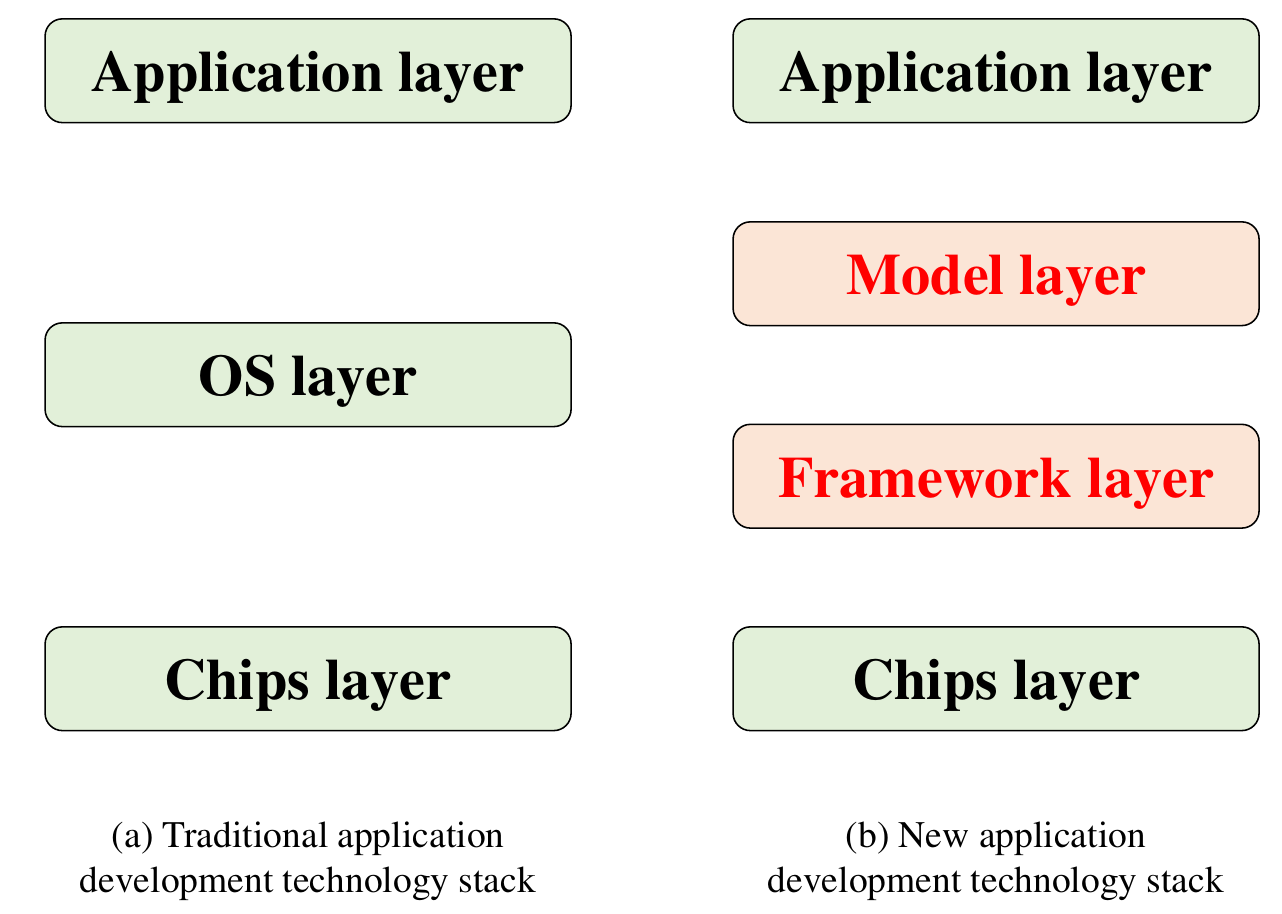}
    \caption{Comparison between traditional and model-based technology stacks.}
    \label{fig:techStack}
\end{figure}

During the whole of Web 2.0, most application development technology stacks can be summarized as a three-tier architecture (Fig. \ref{fig:techStack}(a)). Both desktop and mobile applications depend on their execution environment (i.e., operating systems). As a popular social network and communication platform, there are some obvious differences between the WeChat app \cite{montag2018multipurpose} on Android and iOS systems. For example, on the Android system, the mini-programs can be set as a floating window. Users can quickly open them by sliding to the right on WeChat. Furthermore, users can add mini-programs to the home screen to avoid repeated operations. However, the iOS system does not support the floating window function. If users accidentally close a mini program, they have to open WeChat again and then use the closed mini program. The reason for the above case is that the differences between the two operating systems themselves lead to distinct development environments and design principles. Essentially, this is due to the significant differences in chips used by Android and iOS systems. 

However, these applications will be unified within the MaaS platform. The new application development technology stack is a four-tier architecture (Fig. \ref{fig:techStack}(b)). Compared with the traditional technology stack, the training framework and model layers replace the original operating system layer. Almost all large models are conducted by similar training frameworks (e.g., DeepSpeed \cite{rasley2020deepspeed}, MindSpore \cite{huawei2022huawei}, Megatron \cite{shoeybi2019megatron} Fairscale \cite{baines2021fairscale}, PaddlePaddle \cite{hu2021efficient}), which reduces the effect of operating systems. Besides, the AI-based application is a new kind of product that was developed based on the understanding, generation, and memory capabilities of large models. Indeed, MaaS has gradually become an integral part of the modern technological landscape. In this section, we will explore the related technologies of MaaS. Several technologies and components are involved in the implementation of MaaS. Here are some important elements we suppose:

\textbf{Cloud computing}: MaaS relies on cloud computing infrastructure \cite{hilley2009cloud,khmelevsky2010cloud} to host and deploy machine learning models. Cloud services grant access to scalable and reliable platforms that effectively meet the computational demands of serving models to numerous users. The reliance on cloud computing infrastructure marks a significant shift in how machine learning is utilized. Traditionally, setting up and maintaining the necessary computational resources for machine learning tasks was a cumbersome and expensive process.

\textbf{Model training and optimization}: MaaS providers train machine learning models on large datasets to achieve high accuracy and performance. This involves using techniques like deep learning \cite{rasley2020deepspeed,rajbhandari2020zero}, transfer learning \cite{joy2019flexible,jiang2020individual}, and ensemble learning \cite{kadam2020performance,wang2020ensemble} to develop robust models. Besides, trained models are optimized for deployment on MaaS platforms. This includes techniques like model compression, quantization, and pruning \cite{kim2021pqk}, which reduce the model size and improve its efficiency without sacrificing performance.

\textbf{API and development tools}: API and development tools are key technologies that make sure MaaS services are easy to use. MaaS platforms provide a set of APIs for developers to dumb down the use and integration of models. These APIs provide endpoints for making requests and receiving predictions or insights from the models. Correspondingly, development tools such as software development kits (SDKs), command-line interfaces (CLIs), RESTful APIs, and GraphQL are commonly used in MaaS implementations \cite{nguyen2017exploring,ross2023programmer,li2023skillgpt}.

\textbf{Monitoring and analytics}: MaaS providers implement monitoring and analytics capabilities to track model performance, usage records, and user feedback. This helps in optimizing models, identifying potential issues, and continuously improving the service quality. Analytic tools allow MaaS providers to recognize abnormal decisions, user preferences, and context while running models. The collected information is invaluable for adjusting models to meet the specific needs of users. For instance, if analytic results reveal that users predominantly employ a chatbot for customer support inquiries during peak hours, providers can allocate additional resources to ensure seamless service during these times.

\textbf{Scalability and load balancing}: The essence of scalability in MaaS platforms is all about flexibility and adaptability. In general, the platforms have to handle numerous concurrent requests from users \cite{xu2011intelligent,mesbahi2016load}. When the demand for services surges, a scalable system can timely manage its resources to meet the increased load. This means the system can swiftly allocate more computing power, storage, and bandwidth, ensuring that user experiences remain smooth and uninterrupted. Scalability and load-balancing technologies ensure that the system can handle the increasing demand and distribute the workload efficiently across multiple servers or instances.

\textbf{Security and privacy}: MaaS platforms are confronted with the imperative task of mitigating security and privacy risks inherent to the management of sensitive data and model accessibility. To enhance the integrity of these platforms, a comprehensive arsenal of security measures is deployed, including encompassing data encryption, access control mechanisms, and the utilization of secure communication protocols \cite{shokri2015privacy,papernot2018sok}. Therefore, these strategies collectively serve the dual purpose of safeguarding user data and preserving the sanctity of confidential information.

\section{Advantages of MaaS} \label{sec:Advantages}

Because of its benefits, MaaS is very useful for developers and businesses that are working on making and using GenAI models. MaaS speeds up the entire model development and deployment lifecycle, effectively reducing technical barriers and delivering a series of benefits. These benefits encompass robust performance, flexible payment options, and continuous optimization. Some significant advantages of MaaS are listed as follows:

\textbf{Lower technical barriers}: MaaS effectively lowers the technical barriers associated with the utilization of GenAI models. Developers do not need to become experts in machine learning technologies or master complex algorithms to effectively use advanced models for their development and application needs. They can pay more attention to the creative and practical works. In addition, this kind of user-friendly approach also will encourage innovation across different industries.

\textbf{Simplified model development}: MaaS offers pre-trained AI models as accessible services. Although the pre-train paradigm lets developers leverage AI capabilities without the need for extensive model construction and training, traditional AI model development is still time-consuming and resource-intensive work. Without constructing and training model steps, MaaS empowers developers to seamlessly integrate open-source models into their workflows. This not only optimizes time and resource allocation but also significantly diminishes the initial learning curve and implementation barriers.

\textbf{High performance and scalability}: Since cloud computing services are the basic infrastructure of MaaS, the powerful capability of high-performance computing resources meets numerous data and complex computational requisites. Scalability is an important characteristic of MaaS, which ensures MaaS automatically adjusts resource allocation in response to changing computational requirements. In particular, adaptability is crucial in scenarios where workloads vary dramatically, such as in video production applications. For instance, a model used for real-time image recognition may experience significant spikes in demand during peak usage hours. The cloud infrastructure supports MaaS to seamlessly allocate additional computing resources to handle these spikes and thus ensures fast and consistent response times.

\textbf{Shared knowledge and collaboration}: Since MaaS is built upon a wealth of data and expert experience, the models usually represent broad knowledge and generalization abilities. Developers benefit from this feature without considering gathering or processing large amounts of data independently. The collective knowledge and experience underpinning MaaS models serve as a valuable resource that transcends the confines of individual capabilities. Developers can leverage this shared knowledge to finish their projects as soon as possible. Moreover, MaaS serves as a conducive platform for sharing models and exchanging experiences, which will promote collaboration and knowledge sharing among developers. This accelerates innovation and problem-solving and drives the development of the entire machine-learning community.

\textbf{Intelligence decision support}: MaaS can offer businesses intelligent decision support by leveraging machine learning models to predict future business trends and financial conditions, aiding enterprises in making wiser decisions. MaaS translates analytical results into user-friendly reports and visualizations and provides customized business solutions. Enterprises can select different machine learning models and data processing algorithms based on their specific business requirements and data characteristics, thus achieving more intelligent and personalized business processes \cite{papa2020improving}.

\textbf{Flexible payment models}: In fact, MaaS conventionally utilizes flexible payment models. The subscription-based model allows users to pay only for the actual usage of MaaS services. This kind of payment model is a cost-effective solution and beneficial for small and medium-sized enterprises as well as individuals. Besides, it can convert customers into subscribers, which reduces the cost of attracting new customers and increases customers' loyalty. This is particularly valuable in competitive markets where attracting new customers is expensive and hard \cite{labus2010crm}. Additionally, the subscription model not only ensures a steady income stream for enterprises but also enables providers to deliver consistent updates, improvements, and support services to subscribers.

\textbf{Wide scope of applications}: The GenAI model exhibits compatibility across diverse domains (e.g., natural language processing, image creation, recommendation systems, and video production). Large language models, with their extensive training on diverse data sources, show the ability to perform well in multimodal downstream tasks \cite{driess2023palm,zang2023contextual}. The versatility of GenAI models outperforms the constraints of specific industries. Developers no longer need to find fine-tuned models for different cases. Instead, they can draw from a unified model service that adapts to the unique demands of various application scenarios.

\section{Applications within MaaS} \label{sec:Applications}

\begin{figure}[!ht]
    \centering
    \includegraphics[scale=0.265]{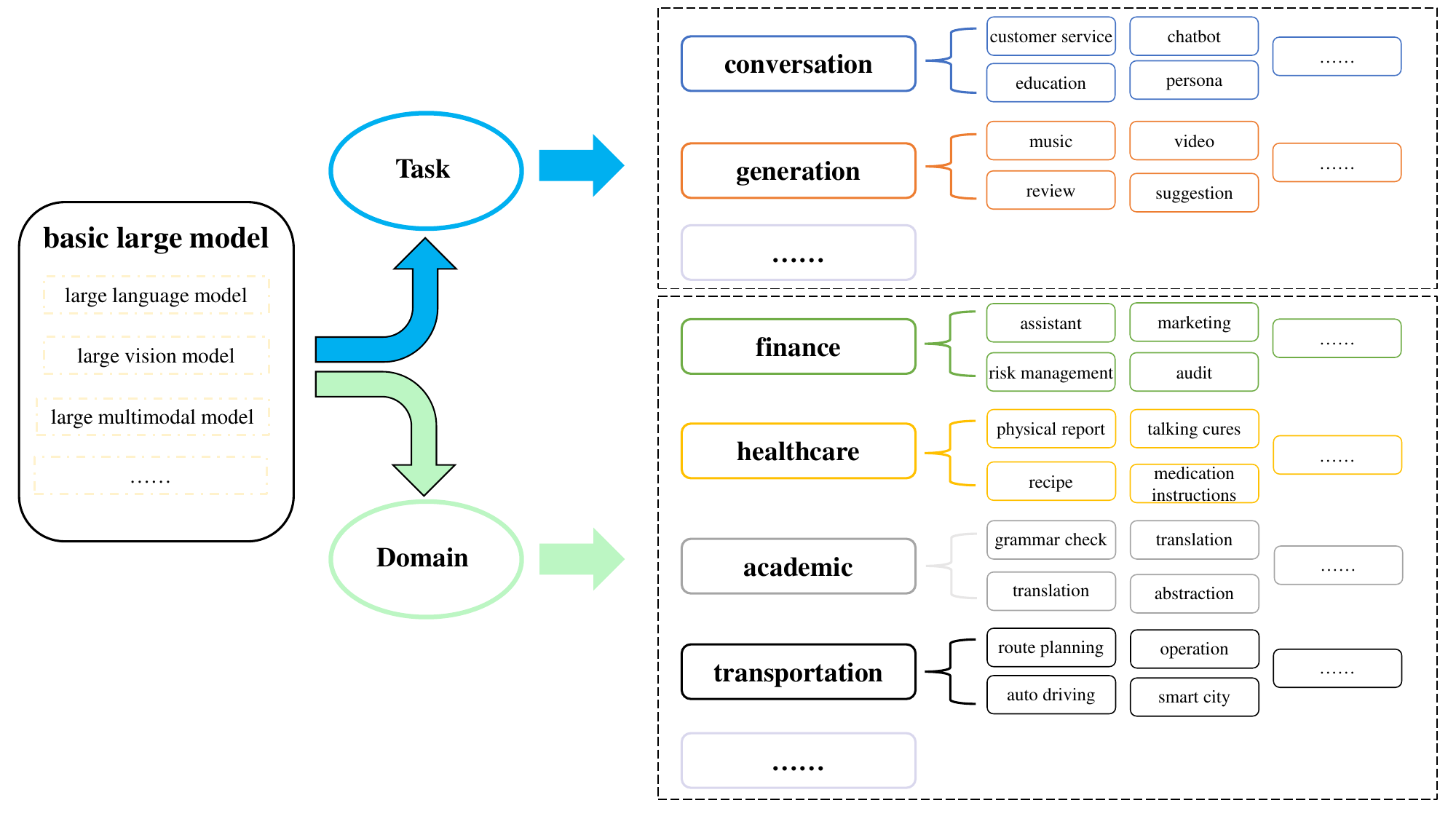}
    \caption{The applications of various industries within MaaS.}
    \label{fig:applications}
\end{figure}

MaaS has a wide range of applications in various domains, including but not limited to the following. Firstly, we present an overview of applications within MaaS in Fig. \ref{fig:applications}.

\subsection{MaaS in Healthcare}

MaaS will have multifaceted implications for the healthcare sector in the future. Its capacity to amalgamate a wealth of diverse healthcare data sources, encompassing electronic health records (EHRs), medical literature, genomics data, and real-time patient monitoring data, is the cornerstone of its transformative potential \cite{kosklin2023knowledge}. MaaS's analytical prowess transcends data integration, enabling it to unveil intricate patterns, correlations, and insights that underpin treatment recommendations and prognosis assessments. The integration of clinical guidelines and principles of evidence-based medicine further distinguishes MaaS as a pivotal healthcare ally. Its analytical engines, fueled by extensive medical literature and clinical trial data, distill the most up-to-date and efficacious treatment modalities for specific medical conditions and patient profiles, thereby furnishing healthcare professionals with an invaluable resource for decision-making. MaaS's capabilities extend to patient profiling and risk assessment, where it leverages machine learning algorithms to scrutinize a series of patient attributes. By juxtaposing this data against expansive patient cohorts and research findings, MaaS crafts personalized risk assessments, forecasting disease progression, treatment responses, and potential adverse events with remarkable precision. As a decision support system, MaaS offers physicians real-time, evidence-based guidance. Through the amalgamation of patient-specific data and clinical knowledge, MaaS optimizes treatment plans, suggesting tailored medication options, dosage adjustments, or alternative therapies based on individual patient characteristics. In addition, predictive modeling is another advantage of MaaS. A potent tool for forecasting treatment outcomes and prognostic assessments. By delving into historical patient data and treatment responses, MaaS identifies predictive factors and forges models that estimate the likelihood of treatment success, disease recurrence, or patient survival. Continuously absorbing fresh patient data, treatment outcomes, and clinical research findings, MaaS updates its algorithms and models to adapt the novel emerging evidence. Finally, the result is a system that continually refines its personalized treatment recommendation tasks and prognostic assessment metrics. This ensures the healthcare system remains adaptable for serving.

\subsection{MaaS in Academic}

MaaS has emerged as a transformative force to profoundly influence research fields and community development through a series of profound effects. The most influential is the acceleration of academic research. MaaS empowers researchers with access to pre-trained models, datasets, and infrastructure, thus streamlining the research and development process. This acceleration is pivotal in allowing researchers to channel their energies directly into their research objectives, without the need for extensive investments of time and resources in model creation and training \cite{feser2023innovation}. As a consequence, research cycles become shorter, fostering rapid advancements and breakthroughs in a myriad of research domains. MaaS also carries the torch of democratization in research. By offering readily available models, tools, and resources, it removes barriers for researchers and offers state-of-the-art models, techniques, and tools that require limited resources or expertise in machine learning. This inclusivity opens the doors for a wide range of researchers to benefit from the progress in their research fields. Facilitating collaboration and knowledge sharing represents another pivotal role of MaaS platforms. These platforms serve as a nexus for researchers to exchange models, datasets, and findings, fostering transparency and nurturing a culture of collaboration. This mutual exchange of knowledge and expertise catalyzes the cross-pollination of ideas, expedites research, and promotes interdisciplinary approaches to problem-solving. MaaS platforms also play a substantial role in reinforcing the reproducibility of research results. By providing access to pre-trained models and standardized datasets, we can rigorously validate and benchmark the results against established standards. This transparency elevates the credibility of research findings and permits equitable comparisons between various approaches, thereby encouraging healthy competition and propelling innovation. Perhaps one of the most consequential effects of MaaS is the swift deployment of research outcomes. MaaS empowers us to seamlessly transition their models and solutions into real-world applications. This capacity serves as a conduit for translating research findings into practical use cases, conferring benefits upon industries, organizations, and society as a whole. We can effectively employ MaaS platforms to bridge the gap between academia and industry.

\subsection{MaaS in Blockchain}

The intersection of MaaS and blockchain technology brings forth a series of transformative effects, profoundly influencing data privacy, model sharing, training, and decentralized model inference within the blockchain ecosystem. The fusion of MaaS and blockchain augments data privacy and security. The inherent attributes of blockchain, including decentralization, immutability, and traceability, play a pivotal role in safeguarding data integrity and privacy. By executing model training and inference on distributed nodes, sensitive data is shielded from the risk of transmission to centralized servers \cite{zyskind2015decentralizing}. This integration not only protects users' data privacy but also improves the overarching system's security. Furthermore, MaaS in a blockchain context can actualize model sharing and marketization \cite{chiu2019blockchain}. The blockchain ledger records the ownership and usage rights of models, thereby establishing a framework for model providers to vend or license their models to fellow users. This ingenious marketization mechanism empowers model developers to enhance their creations while simultaneously affording users a wider array of choices and greater flexibility in model selection. In terms of model training, blockchain's verifiability attribute finds its application by validating the credibility of model training processes. This is accomplished by documenting the entire training process and its outcomes on the blockchain, guaranteeing transparency, auditability, and resistance to tampering. This feature is of paramount significance in applications where the trustworthiness and traceability of the model training process are crucial, such as in the domains of finance and healthcare. The decentralized nature of blockchain is further leveraged for decentralized model inference. Models are deployed across multiple nodes within the blockchain, enabling parallel inference and consensus verification of results. This decentralized model inference approach augments both the efficiency and reliability of model inference. It is particularly pertinent in applications requiring high reliability and resilience against tampering, such as in data encryption, smart contract execution \cite{atzei2017survey}, and identity validation and authentication \cite{norta2019safeguarding}.

\subsection{MaaS in Web 3.0}

Web 3.0 will be the next evolution of the Internet, characterized by decentralization, user control, intelligence, and data privacy \cite{gan2023web}. Intelligence is one of the significant features of Web 3.0, as well as of MaaS \cite{alabdulwahhab2018web}. Consequently, MaaS serves as an intelligent solution that smoothly aligns with the ideals of Web 3.0. Besides, Web 3.0 also emphasizes personal data autonomization within user control. Since the individuals' data is a cash cow, in the context of MaaS, users transmit their data to the cloud computing platform for processing while retaining full control over the utilization and dissemination of their data. The flexibility to selectively share data in exchange for model predictions safeguards personal privacy, offering users the autonomy to manage and leverage their data in harmony with the principles of Web 3.0. Data transmission and storage are inherent to the process of sending data to the cloud for model inference. To uphold data privacy and security, MaaS incorporates measures such as encryption, authentication, and access control (as illustrated in the last subsection). These safeguards protect user data from unauthorized access and misuse, in alignment with the data privacy and security objectives of Web 3.0. Web 3.0 also contains smart contract technology, which ensures automated business logic in decentralized applications. By uniting smart contracts with MaaS, a higher level of automated decision-making and intelligent business logic can be achieved. Web 3.0 encourages data interoperability and cross-platform interaction. MaaS complements this objective by offering standardized interfaces and protocols, enabling diverse platforms and applications to effortlessly invoke machine learning models.

\section{Challenges and Issues} \label{sec:Challenges}

Model-as-a-Service (MaaS), as a new deployment and service paradigm for different AI-based models, will face several challenges and potential issues for future development. In this section, we highlight the key challenges and issues associated with MaaS. Details are described below.

First and foremost, one significant challenge that MaaS encounters is ensuring security and privacy. In the architecture of MaaS, models and rich data often need to be transmitted and processed across different environments. This can potentially lead to the exposure of sensitive data and the misappropriation of confidential models. Therefore, ensuring the security and privacy of models and data becomes one of the key issues that MaaS needs to address. Future directions could include the application of encryption and secure computing technologies, along with the establishment of more rigorous access control and permission management mechanisms.

Secondly, MaaS also grapples with challenges related to model management and version control. With models undergoing continuous iterations and updates, effective management and control of model versions have become a critical concern. Additionally, the deployment and service aspects of models must consider scenarios where multiple models coexist. Issues such as model selection, composition, management, and scheduling of different model versions need to be addressed. Future directions might involve the creation of comprehensive model repositories and version control systems, as well as automated model selection and composition algorithms.

Furthermore, MaaS needs to address challenges related to computational power and resource allocation. As machine learning models grow in complexity and scale, the demand for computational resources also increases. Within the MaaS architecture, achieving elastic scalability and load balancing to meet the varying scales and demands of model services becomes a significant challenge. Future directions could include leveraging containerization and cluster management technologies to achieve flexible deployment and scheduling of models \cite{bentaleb2022containerization}, along with optimizing the computational and inference processes of models to enhance resource utilization.

In practice, the future various development of MaaS also focuses on multi-platforms and cross-organizational integration. Deploying and utilizing models often involve collaboration among multiple platforms and organizations. Challenges include achieving model interaction and communication between different platforms and ensuring data sharing and cooperation among various organizations. Future directions might encompass the formulation of standardized model interfaces and communication protocols to facilitate interoperability and collaboration between different platforms.

There are several respects in which MaaS may have trouble:

\begin{itemize}
    \item \textbf{Latency}: Latency is the time between a request and a response. Depending on the complexity of each GenAI model and the volume of requests, a latency issue is inevitable. Real-time applications may require low-latency responses, which may be difficult to achieve, particularly for foundation models. In fact, we need low-latency, real-time inference services for applications. 

    \item \textbf{Interpretability of model and results}: A ``black box" model has poor interpretability, and the users cannot accurately fill in the decision-making results. Almost all GenAI models exhibit a high degree of complexity and perform poorly in their interpretability.  This often poses a significant constraint when it comes to elucidating the rationale behind the models' decisions to users.

    \item \textbf{Governance, risk, and compliance (GRC)}: The main compliance refers to data compliance, content compliance, platform operation compliance, platform management compliance, etc. Organizations often have to comply with various rules and regulations, including but not limited to General data protection regulation (GDPR) \cite{regulation2018general} and industry-specific standards. These guideline steps will meet some significant challenges when we deploy MaaS. 

    \item \textbf{Data quality and bias}: High-quality big data is the cornerstone of foundation models. If the training data used to build models is biased or of poor quality, it can lead to biased predictions and inaccurate results when deployed as a service. Pre-trained large models may inherit biases present in the training data, leading to biased or unfair outcomes \cite{kim2019learning,su2023fake}. Careful evaluation and mitigation of biases are necessary to ensure the ethical and fair application of the models.

    \item \textbf{Lack of transparency}: Ensuring that models can work seamlessly with a variety of platforms and technologies used by different clients can be challenging. Compatibility issues may arise. Pre-trained models used in MaaS may be complex and difficult to interpret. Understanding the inner workings and making decisions based on the model's output can be challenging \cite{jarrahi2023artificial}, particularly for critical applications requiring explainable AI.

    \item \textbf{Dependency on service availability}: Using MaaS from a particular provider can lead to vendor lock-in, making it difficult to switch to a different service or bring the model in-house. MaaS relies on the availability and reliability of the service provider's infrastructure. If the provider experiences downtime or interruptions, it can disrupt the functioning of applications relying on the models.

    \item \textbf{Limited offline capabilities}: Ensuring the availability and reliability of the MaaS can be challenging. Downtime or service interruptions can disrupt applications that depend on the model. MaaS often requires an active Internet connection to access and utilize the models. In scenarios where Internet connectivity is limited or unreliable, offline functionality may be compromised.

    \item \textbf{Limited customization}: This may seem counterintuitive, but it is indeed the case. Sometimes, the vast majority of ordinary users lack the capability to modify these parameters. Some MaaS providers might limit the level of customization available, which can be a limitation for applications with specific requirements. Pre-trained models provided by MaaS may lack the flexibility to be extensively customized according to specific requirements. Users may have limited control over the underlying model architecture and parameters.
\end{itemize}

With MaaS, how to achieve ``one model to serve all"? This is still an open question. Addressing these challenges and limitations often requires careful planning, architecture design, and ongoing maintenance. Organizations should evaluate their specific needs and constraints before deciding to implement MaaS to ensure it aligns with their goals and resources.

\section{Future Directions} \label{sec:Directions}

MaaS holds great potential in the fields of machine learning and artificial intelligence. There are several noteworthy future directions that can take further research. These include, but are not limited to, addressing security and privacy concerns, exploring encryption and secure computing techniques, and developing access control mechanisms. Enhancing model management, version control systems, and optimizing resource allocation for scalability are also the keys to MaaS research. Additionally, cross-organizational integration through standardized interfaces and communication protocols is promising for future work. These directions are pivotal for unlocking the full potential of MaaS. We discuss some of them as follows:

\begin{itemize}
    \item \textbf{Data privacy and security}: With the increasing adoption of MaaS, the growing demand for models and data will raise people's concerns about security and privacy \cite{regulation2018general,chen2022federated}. In the future, there will be a necessity for more advanced encryption schemes, access control mechanisms, and data protection methods to ensure the security of data transmission and processing in MaaS.

    \item \textbf{Algorithm optimization and model efficiency}: Future research can focus on improving algorithms and optimizing model efficiency to enhance the performance and responsiveness of MaaS. The efficient utilization of limited resources is an urgent concern within MaaS. We should explore ways to optimize the computation and inference processes of large models to reduce resource consumption \cite{yang2023large}. This optimization work is vital for maximizing the potential of MaaS in the future.

    \item \textbf{Model lifecycle management}: Model lifecycle management, which covers development, training, deployment, updates, and retirement, is a critical area for research and development. Scholars and engineers can focus on establishing more sophisticated model management theories and developing corresponding version control systems to ensure effective model governance and maintenance.

    \item \textbf{Explainable AI and ethical considerations}: How to improve the explainability of MaaS models and address ethical and moral issues related to AI to ensure fairness and transparency in their applications? In MaaS research, a crucial challenge is the complexity of model decision processes. We must devise novel paradigms to enhance the interpretability of these models and then let users understand the rationale behind the models' decisions.

    \item \textbf{Environmental friendliness and energy efficiency}: Future research can explore ways to improve the environmental friendliness and energy efficiency of MaaS by optimizing the use of computational and communication resources, reducing its impact on the environment.

    \item \textbf{Multimodal learning and cross-device collaboration}: We can explore multimodal learning \cite{wu2023multimodal} (involving multiple modalities) and cross-device collaboration (e.g., federated learning) to enable the sharing and training of models in distributed environments, improving model performance while preserving data privacy.

    \item \textbf{Social impact and policy issues}: We must carefully consider the societal implications of MaaS and actively contribute to the establishment of regulations and policies that will guide its utilization and advancement. It is imperative that these regulations align with ethical principles and the broader interests of society.
\end{itemize}

\section{Conclusion} \label{sec:Conclusion}

This is the first survey that concentrates on the topic of MaaS based on GenAI models. We first discuss the history between XaaS models, which we suppose is a continuous conversion. Then, we review the relationship between cloud platforms (i.e., SaaS, PaaS, and IaaS) and draw out differences from the perspective of Web eras. Several important technologies of the MaaS platform are introduced in detail, as well as their benefits. Subsequently, we present some application scenarios, since MaaS has a wide range of applications in various domains. Finally, we identify several key problems and challenges that MaaS will face and point out some future directions for studying MaaS. MaaS has emerged as a transformative paradigm in the era of AI, enabling organizations to leverage pre-trained models seamlessly. Its benefits, including accelerated adoption, low cost, and easy operation advantages, have paved the way for widespread AI integration in industries. As MaaS grows in the future, data privacy, customization, and ethical considerations will be key troubles and challenges to taking full advantage of its potential advantages. With the evolution of MaaS for social good, ubiquitous AI technologies will be deeply engaged in our daily lives.

\section*{Acknowledgment}

This research was supported in part by the National Natural Science Foundation of China (Nos. 62272196 and 62002136), the Natural Science Foundation of Guangdong Province (No. 2022A1515011861), and the Young Scholar Program of Pazhou Lab (No. PZL2021KF0023). 

\bibliographystyle{IEEEtran}
\bibliography{paper}

\end{document}